\documentclass[conference]{IEEEtran}
\usepackage{amsmath}
\usepackage{amssymb}
\usepackage{amsthm}
\usepackage[dvipsnames]{xcolor}
\usepackage{graphicx}
\usepackage{booktabs}
\usepackage{makecell}
\usepackage{comment}
\begin{document}
%
\title{ Evaluating complexity and resilience trade-offs in emerging memory inference machines }

\author{\IEEEauthorblockN{Christopher H. Bennett*, Ryan Dellana*, T. Patrick Xiao, Ben Feinberg, Sapan Agarwal, \\
 Suma Cardwell, Matthew J. Marinella,  William Severa, Brad Aimone \\}
\IEEEauthorblockA{Center for Computing Research, Sandia National Laboratories, Albuquerque, New Mexico\\
Email: \{ cbennet, jbaimon \} @sandia.gov \\
*These authors contributed equally}}

\maketitle

\begin{abstract}
Neuromorphic-style inference only works well if limited hardware resources are  maximized properly, e.g. accuracy continues to scale with parameters and complexity in the face of  potential disturbance. In this work, we use realistic crossbar simulations to highlight that compact implementations of  deep neural networks  are unexpectedly  susceptibilie to collapse from multiple system disturbances.  Our work proposes a middle path towards high performance and strong resilience utilizing the \textit{Mosaics} framework, and specifically by re-using  synaptic connections in a recurrent neural network implementation that possesses a natural form of noise-immunity.
\end{abstract}


%
\IEEEpeerreviewmaketitle

\section{Introduction}
A variety of neuromorphic systems have been implemented using emerging memory devices, but  few perform at industrial levels  due to the difficulties of implementing software-quality synapses and gradients. These challenges are somewhat simplified in the inference-only application, where imported weights from a separately pre-trained network are used to accelerate predictions on a known task. Nevertheless, electrical issues such as parasitics, cycle-to-cycle read noise, and device-to-device variability may limit the ultimate size and accuracy of inference-only neuromorphic accelerators \cite{agarwal2016resistive,bavandpour2018mixed}. These issues can quickly degrade the performance of deep networks with hundreds of thousands of parameters.  Accelerators have been proposed to leverage memristive crossbars such as PUMA \cite{puma}, ISAAC, \cite{bojnordi2016memristive}, but they typically discuss the implications of these issues at minimal length. We explore the idea that these systems may be unexpectedly fragile to combined perturbations, and seek mitigations.

\begin{figure}[ht]
	\centering
	\includegraphics[width=1.8in]{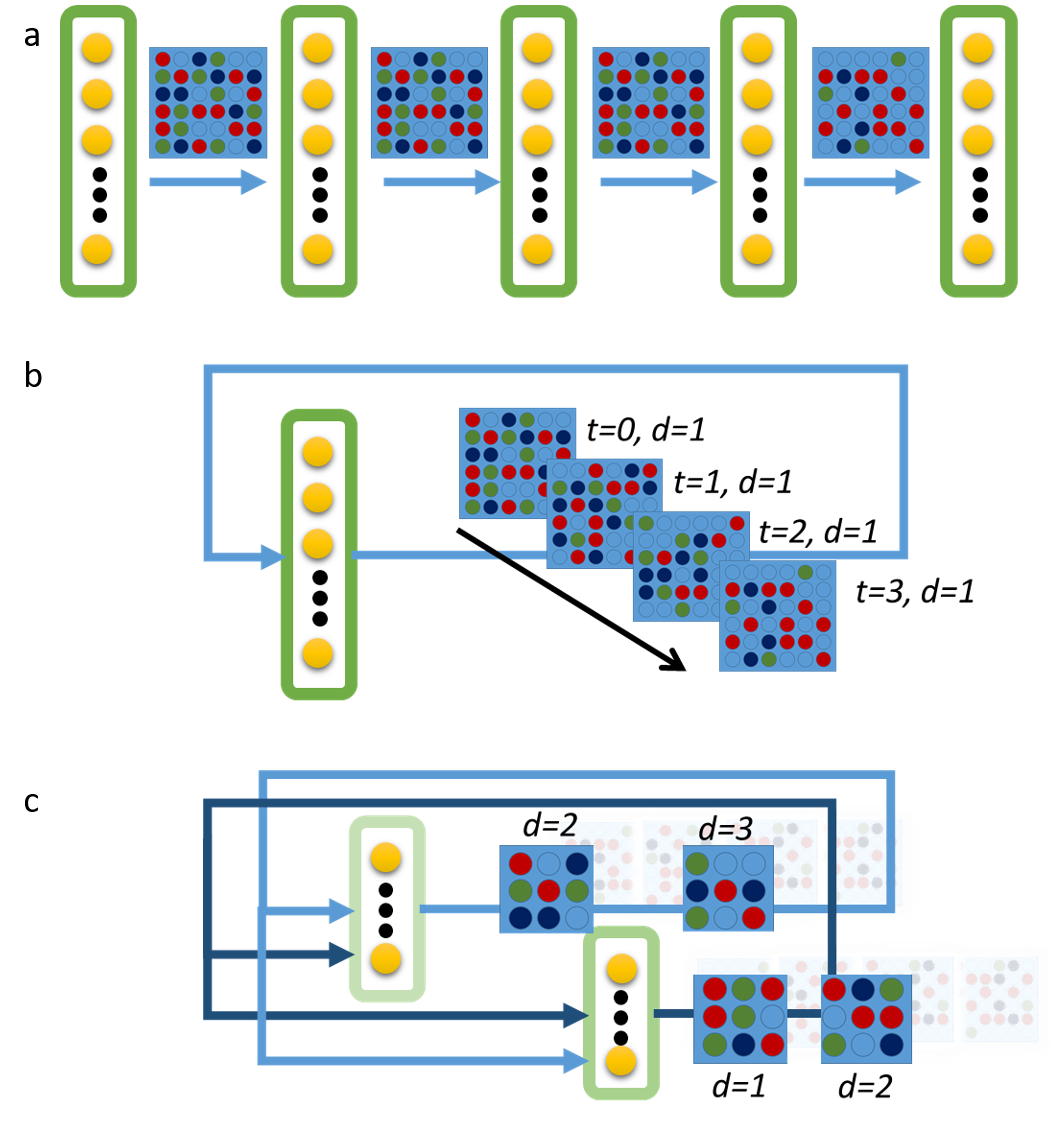}
	\caption{(\textbf{a}) Simple ANN depiction with yellow sub-graphs as neurons and blue tiles as weights. (\textbf{b}) A recurrent Mosaic implementation, with processing neurons recurrently connected to a dynamically used single set of weights. (\textbf{c}) Future \textit{Mosaics} implementation, with each layer broken into component sub-layers.  This requires that the connection matrices be partitioned with different delays to permit connections to `skip' layers to target appropriate node}
	\label{MosaicFigure}
\end{figure}

Neural networks with an intrinsic temporal behavior  may yield new sources of power and resilience \cite{aimone2019neural}. In order to efficiently map standard networks to temporal ones, the efficiency of fundamental operations in a given graph must be considered. As illustrated in Fig. \ref{MosaicFigure}, while the overall size of an artificial neural network (ANN) graph may be very large, parts of the necessary computation are done repeatedly. \textit{Mosaics} in this context refers to a temporal form of neuromorphic multiplexing, whereby certain computations (in this example, the neurons, which are often a limiting resource) can be reused for different stages of an algorithm; allowing larger scale algorithms to be implemented on a resource restricted neuromorphic platform.   By exploring the partitioning of a neural graph into sub-graphs that enable the computation to be performed on a smaller subset of available computing nodes - trading space for time - we open a new avenue for failure and resilience analysis. Concretely, in the following sections we explore this contrast through three relatively well-known ANNs: a) a simple, low parameter multi-layer perceptron, b) a complex, medium parameter convolutional neural network (CNN), and c) a medium-parameter, medium complexity recurrent neural network (RNN) inspired by \textit{Mosaics} re-use concepts. In general, RNNs are an attractive emerging option for the demonstration of on-chip learning or inference, as they are powerful general computational models \cite{hammer2000approximation} . Recently, long-short-term memory (LSTM) networks have been the most heavily considered for implementation   with dense non-volatile memory arrays \cite{gokmen2018training}; however, such schemes involve complex crossbar partioning and hardware implementation of many transcendental functions (e.g. $tanh$) on the edge of the periphery. Feasibly, LSTM implementations can be done on-chip during inference mode, such as with phase-change memory (PCM) NVM synapses \cite{tsai2019inference}, yet the full overhead required to implement such systems has not been documented. Drawing on previous work which shows that an RNN network using appropriate, simple (ReLU) activation functions can still perform competitively to an LSTM on certain tasks \cite{le2015simple},  in the following sections we demonstrate that a vanilla RNN system can perform competitively to a compact CNN on two state of the art machine learning (ML) tasks, at a lower energy budget. 

\begin{figure}[ht]
	\centering
	\includegraphics[width=2.75in]{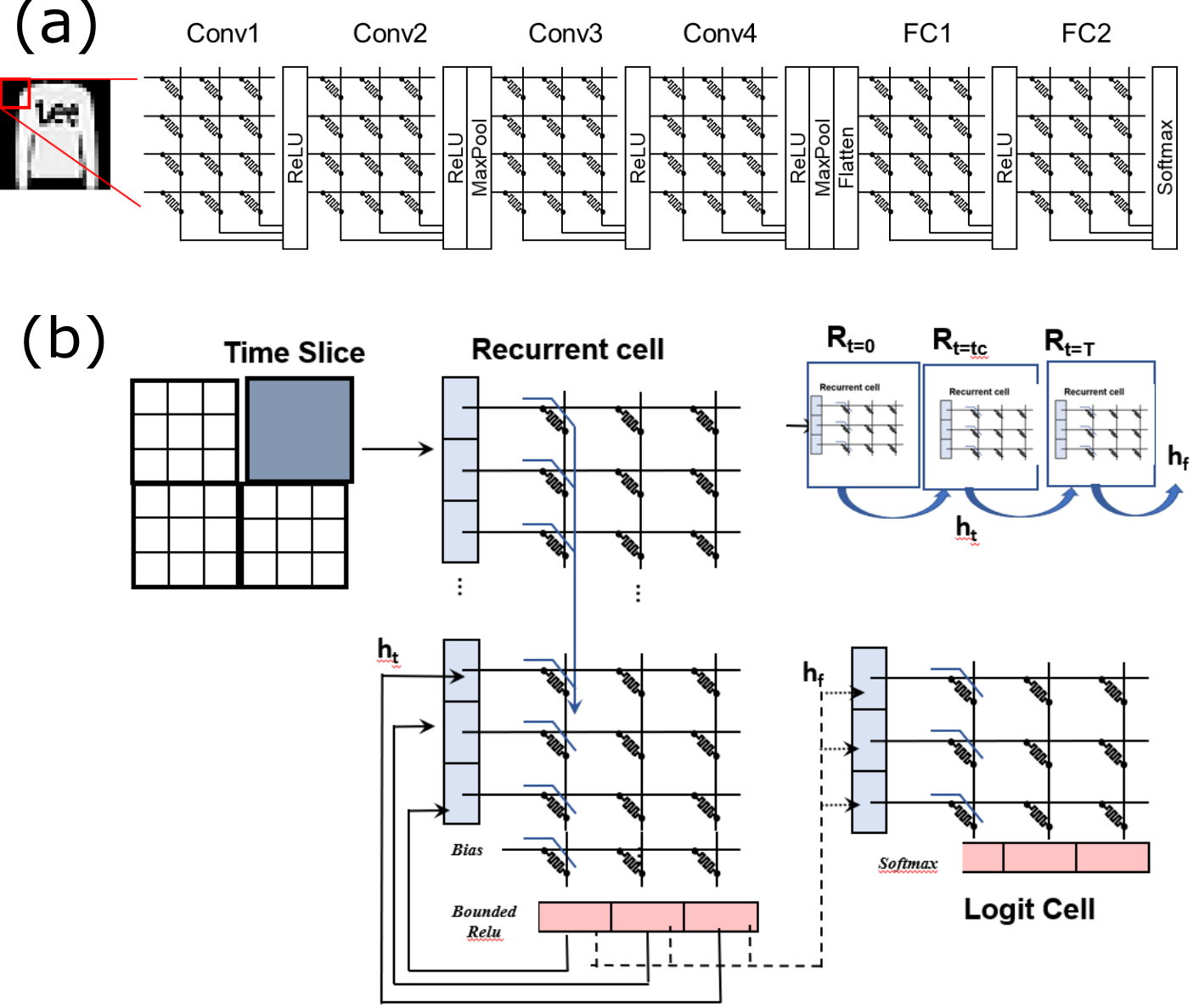}
	\caption{(\textbf{a}) A CNN architecture is demonstrated;  parts of images are fed to ANN sequentially.  (\textbf{b}) As in the CNN, pieces of the image are fed to the recurrent cell until the entire image has been presented; now, the read-out or fully connected logit system only activates on the last time-step.}
\end{figure}

\section{Methodology}
We trained CNN models using the Keras framework \cite{chollet2015keras} on the iconic  MNIST data task \cite{lecun2010mnist} as well as the newer fashion-MNIST data task \cite{xiao2017fashion}.  We then imported these models into CrossSim, a crossbar simulation tool that helps model resistive memory crossbars as highly parameterizable neural cores \cite{CrossSim}, and which has recently been extended to  perform physics-rich analysis of inference operations. Our imported models for CNNs contain 119,322 trainable parameters and contain both convolutional and fully connected (FC) layers, each followed by a bounded rectified linear function (ReLU, given as $f(x) = min(max(0,x),1)$) as visible in Fig. 2(a); exact model is given in Table 1.  \cite{bennettSONOS} and as visible in Fig. 2(a). Our imported MLP models are standard consisting of one hidden layer (128 ReLU functions) and a logit of 10; this model has 101,770 trainable parameters. Our recurrent neural network design, newly developed for this work and visible in Fig. 2(b), consists of a recurrent matrix which is re-used during multiple time steps. Notably it is partioned into two cores, a core which always receives a component of the input and a part which receives the hidden layer activation $h_t$ from the previous time step $t-1$ (or null if $t=0$) . The number of output nodes  $D_{o}$ is set at 400, and depending on the chosen time steps $t$, the number of hidden nodes $D_{hl}$ receiving activation from the previous time step is given as $301-M/t$ , where $M=784$ is the dimensionality of our task. The number of parameters of our RNN networks vary from 118,37 trainable parameters at $t=7$ up to 158,656 parameters at $t=56$ ( time step integers are divisors of $M$). While the the number of physical weights does not increase with the number of time-steps in CrossSim, the number of parameters tracked in Keras tracked for backprogagation of error does.

 The RNN core crossbar  is connected to a read-out/logit crossbar of dimensions $D_{o}$ x $L$, where $L$ is the number of classes (here $L=10$); critically, the second core  is only activated at the final time step (every $t$ steps).  In our resilience testing strategy, we have considered two classes of pre-trained networks: noise-prepared or regularized networks,  which train with  jittered gaussian filters on every hidden neuron's ReLu function during training following the scheme given in \cite{bennettSONOS} and standard/un-prepared networks which have been trained without noise. As first suggested in \cite{noh2017regularizing}, the use of noise regularization provides a definitive improvement in inference (Test ) performance of models to internal and external noise or perturbation effects. During training phase, noise centered around 0 with a distribution width $\sigma_\text{neu}$ is injected into every ReLu neuron; during testing phase  $\sigma_\text{syn}$ is then injected on a synapse-per-synase basis . This device-by-device synaptic dispersion most closely relates to  intrinsic noise effects that would occur during  the operation of large arrays (Johnson-Nyquist noise), but can approximate shot noise or device-specific perturbations or weight variance as suggested in \cite{agarwal2016resistive}. In addition, given the test set  $\mathcal{Y}$, we have also considered the addition of additive gaussian noise at dispersion   $\sigma_\text{te}$ where $\mathcal{Y}_\text{noisy}  = \mathcal{Y} + y(\sigma_\text{te})$.  We have perturbed all weight matrices in every system consistently based on the generation of random numbers with seeds different in every run.

\begin{table}
   \caption{Primary System Architectures.} 
   \label{tab:sys}
   \small 
   \centering 
   \begin{tabular}{lccr} 
   \toprule[\heavyrulewidth]\toprule[\heavyrulewidth]
   \textbf{Neural Net Design} & \textbf{Layout}  \\ 
   \midrule
   MLP & 785x300, 301x10 \\
   RNN & 301x400, ($t$ time steps), 401x10  \\
   CNN & C3/3-C3/3-MP2-C3/6-C3/6-D100-D10 \\
   \bottomrule[\heavyrulewidth] 
   \end{tabular}
\end{table}

\begin{table}
	\caption{Standard Resilience results ($t=7$)}
	\label{Table}
	\centering
	\begin{tabular}{@{}l*{4}{c}@{}}
		\toprule
		\text{Architecture} & \multicolumn{4}{c@{}}{\text{Noise Scenario}}\\
		\cmidrule(l){2-5}
		& Internal ($\sigma_\text{syn}$*)  & External  ($\sigma_\text{te}$*) &Both Effects \\
		\midrule
		\makecell[l]{MLP- MNIST} &  $96.8 \%$ & $94.1 \%$ & $93.1 \%$ &\\
		\makecell[l]{RNN - MNIST} & $97.4 \%$ & $95.1 \%$ & $94.9 \%$ & \\
		\makecell[l]{CNN-MNIST} &  $98.5 \%$ & $96.7 \%$ & $96.05 \%$  &\\
		\midrule
		\makecell[l]{MLP- f-MNIST} & $82.2 \%$ & $69.91 \%$ & $62.35 \%$ & \\
		\makecell[l]{RNN - f-MNIST} & $86.3 \%$ & $84.22 \%$ & $81.11 \%$ &  \\
		\makecell[l]{CNN-f-MNIST*} & $85.1 \%$ & $57.91 \%$ & $42.35 \%$ &  \\
		\bottomrule
	\end{tabular}
	\\
	{\raggedright *In all cases, $\sigma_\text{syn} =\sigma_\text{te} = 0.025$ \par}
\end{table}

\section{Results}
\subsection{Resilience to noise sources}
First, we consider the raw resilience of MLP, RNN, and CNN networks to noise at physically plausible values. Noiseless  NVM inference performance values for the MLP, RNN and CNN sit at $97.7 \%, 98.5 \%$, and  $99.1 \%$ for MNIST, and at $84.5 \%, 89.1 \%$, and$ 90.1 \%$ for f-MNIST.  As demonstrated in Table II, all considered systems degrade least from these results given only internal noise, second best with just external noise, and suffer the most when the effects are compounded. There is a significant task-dependence, with CNN systems performing best on MNIST (easier task) even in the worst-case, and with RNN systems performing by far the best on the fashion-MNIST (harder task) case. Of particular concern/interest is the degradation of the CNN f-MNIST models to test-set noise degradation and combined noise degradation, which does even worse than the MLP system. Although  the possibility of catastrophic responses of CNN networks to adversarial perturbations are welll-known, this effect can probably be at least partially lessened by a more complex CNN architecture with larger filter sizes \cite{yang2019design}. 

Next, we consider broader sweeps of the internal noise parameter  $\sigma_\text{syn}$ along with a consideration of the effect of noise-regularization during inference time for the two best-performing systems overall (CNN, RNN). As visible in Figure 3, in both the MNIST (a,b) and f-MNIST (c,d) cases, the stochastic ReLU behavior during training helps the CNN systems far more than the RNN systems. For MNIST, the regularization helps the CNN achieve best overall performance at both the purely internal and combined noise source cases (blue improves to orange), while the regularization assists the RNN but not as dramatically (green to red). This same trend holds in fashion MNIST with only internal noise; however, in the combined noise case for fashion-MNIST, at $\sigma_\text{syn}<0.075$ the regularized and non-regularized models are broadly superior to CNN approaches. Overall, the current results support the argument that RNNs seems to have greater intrinsic noise immunity as compared to CNNs, while CNNs benefit far more from a standard neuron-level regularization approach.

\begin{figure}[ht]
	\centering
	\includegraphics[width=3.1in]{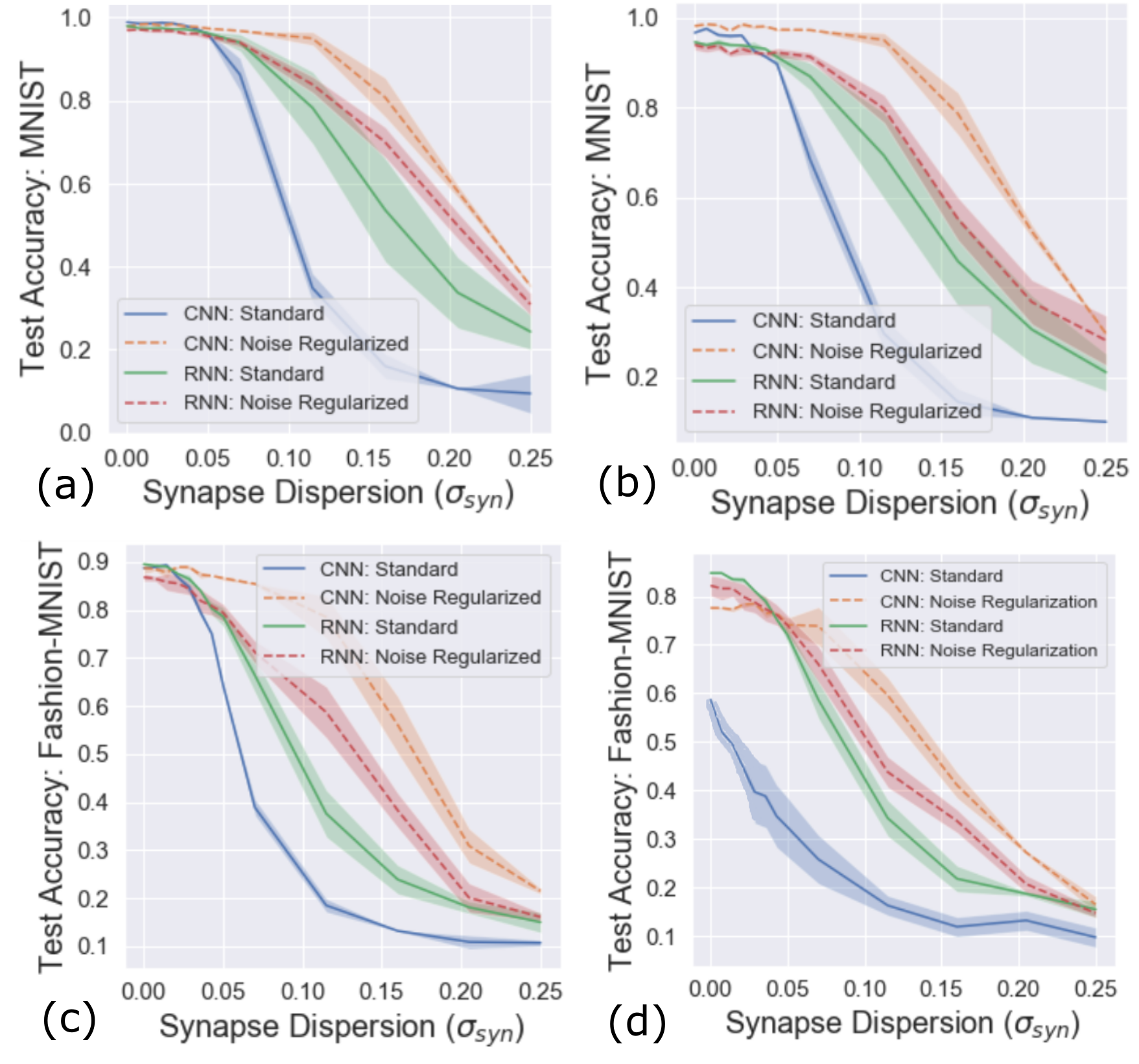}
	\caption{(\textbf{a}) Test inference accuracy on the MNIST task for regularized and non-regularized CNN and RNN systems as a  function of progressively increased   $\sigma_\text{syn}$ value; (\textbf{b})) MNIST performance on highlighted systems given both internal and noisy test set scenario ( $\sigma_\text{te}=0.025$). (\textbf{c}) fashion-MNIST performance given internal noise and (\textbf{d}) both internal and test-set noise fashion-MNIST degradation (again $\sigma_\text{te}=0.025$).}
\end{figure}

\begin{figure}[ht]
	\centering
	\includegraphics[width=3.25in]{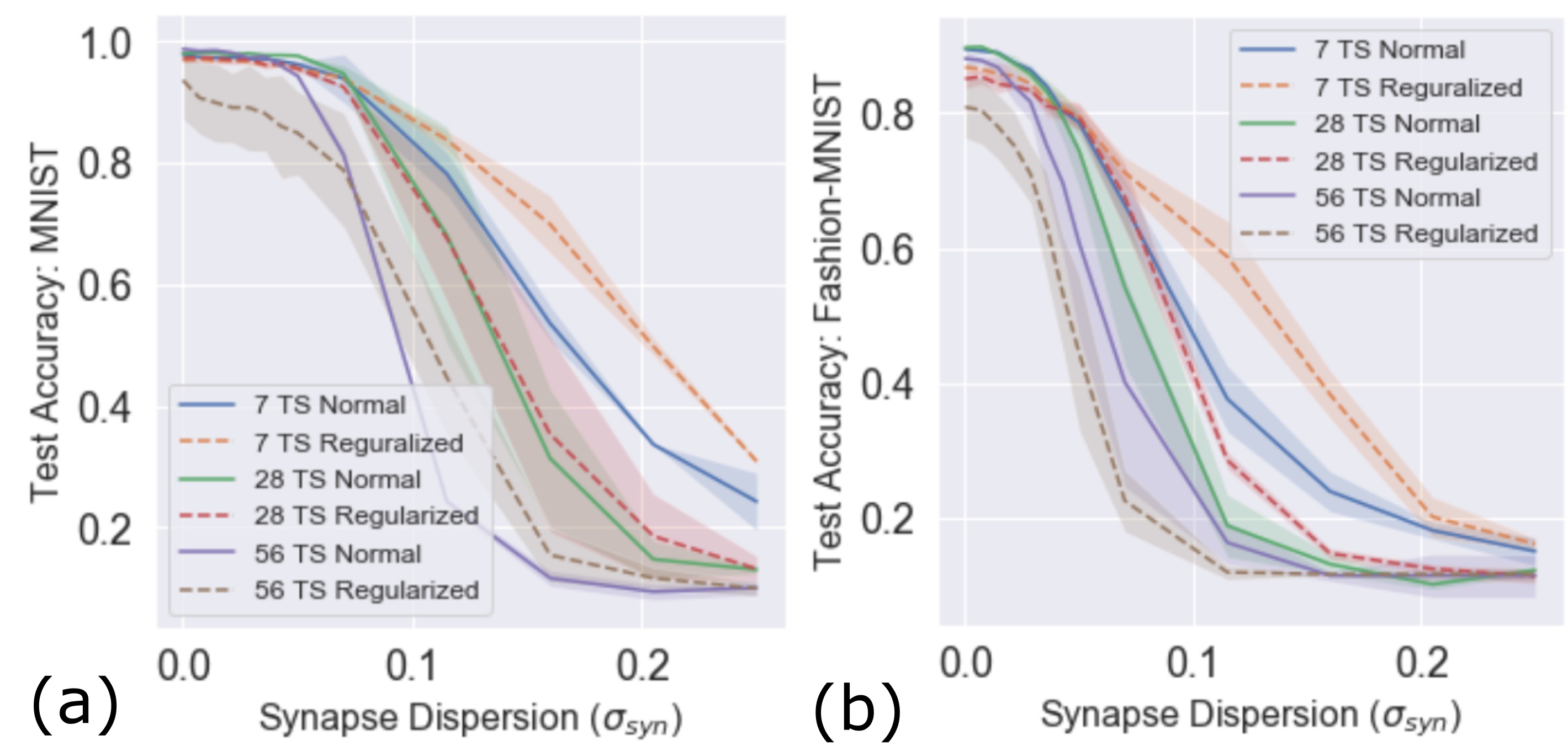}
	\caption{(\textbf{a}) Test inference accuracy on the MNIST task as a function of the internally applied perturbation parameter for all of the considered time-step values , in both regularized and non-regularized procedures (\textbf{b}) Same analysis is presented for all pre-trained models on the fashion-MNIST task.}
\end{figure}

\subsection{Temporal stacking sensitivity of RNN}
The considered RNN approach can treat the number of time-steps of input presentation , which are automatically concatenated at the output of the RNN core, as a free parameter. As noted in \cite{le2015simple}, extending $t$ too far can cause problems in learning convergence since the temporal dependencies (trace) between the eventual output and input stream can become too weak. We find a similar result but with even greater sensitivity in the NVM system. As shown in Figure 4,  when $t=26$ a slight degradation is visible, and at $t=56$ inference at the task even at no noise  loses $8-10 \%$ depending on the task.

\subsection{Energy overhead considerations}
We combine elementary energy costs and system dimensions to estimate energy overhead. Two primary energy costs drive inference: the vector-matrix-multiply (VMM) operation, which requires signficant energy to charge and read out (given ADC costs) the memristive array, and the neuron activation energy costs. Using the methodology proposed in \cite{marinella2018multiscale}, the CMOS ReLU circuit given in \cite{priyanka2018cmos}, and considering two device candidates for inference - optimized filamentary resistive RAM (ReRAM) alongside charge-trap based ultra-low energy SONOS memory \cite{agarwal2019designing}- we find  that emerging memory CNN systems are energy hungry relative to the other options (Table III). Our proposed RNN system consumes only 7x more than the MLP system by exploiting the weight re-use or temporal encoding strategy, and saves between 13-38x energy compared to the competitor CNN system.
\begin{table}
	\caption{Energy overhead per inference op in considered systems}
	\label{Table}
	\centering
	\begin{tabular}{@{}l*{4}{c}@{}}
		\toprule
		\text{Noise Mode} & \multicolumn{4}{c@{}}{\text{Synapse Type}}\\
		\cmidrule(l){2-5}
		& Total Energy/Op & VMM Op & Neuron Activation Op\\
		\midrule
		\underline{MLP ReRAM*} & 4.24 nJ  & 4.22nJ & 15pJ & \\
		\makecell[l]{RNN ReRAM* $\dagger $} & 35.6nJ & 35.5nJ & 66pJ & \\
		\makecell[l]{CNN ReRAM*} & 480 nJ & 479 nJ &  358 pJ  &\\
		\midrule
		\underline{MLP SONOS*} & 6.04 nJ  & 6.02nJ & 15pJ  \\
		\makecell[l]{RNN SONOS* $\dagger$} & 42.7nJ & 42.7nJ & 66pJ \\
		\makecell[l]{CNN SONOS*} & 2.084 $\mu J$ & 2.084$\mu J$ &358 pJ &  \\
		\bottomrule
	\end{tabular}
	\\
	{\raggedright *We have assumed that softmax (all systems) and maxpool ( CNN only) operations energy costs are negligible. \par}
	{\raggedright $\dagger $RNN energy estimates are worst-case ($t=7$). Note benefits are greater for greater number of $t$. \par}
\end{table}

\section{Discussion}
One phenomenon suggested but not proven in our analysis is the idea that noise regularization is still helpful , but less critical than in more complex CNN models, due to the presence of attractor basins in recurrent architectures which help to provide some intrinsic level of noise resilience \cite{ceni2019interpreting}. We have also discovered that binary stochastic neurons may not be sufficient for full RNN regularization; a more complex function or method may be required. One interesting method suggested in \cite{campos2017skip} would be the use of temporal skip connections. Another fruitful extension of this work could be the exploration of how recurrent ,  convolutional , or mixed architectures using effective regularization mtehods (tuned based on the time-step dynamics of networks) could provide defensive properties against adversarial noise or perturbation effects  \cite{you2019adversarial,jin2015robust}. Finally, while we have showed the \textit{Mosaics} concept implemented in the temporal domain in this work, the efficient parallel or horizontal stacking of convolution operations may yield more efficient and/or  resilient convolutional operations (Figure \ref{MosaicFigure}c).

\section{Conclusion} 
Making emerging memory inference systems more reslient is an important goal for the neuromorphic engineering field, but so far analysis has focused almost exclusively on the limitations of CNN systems. In this work, we have put CNN resilience in conversation with other approaches and in particular our novel RNN approach. We have shown that an ostensibly simple recurrent neural network design  efficiently implements the idea of temporal stacking or weight re-use and reduces energy costs by at least 13.5x while achieving results that are competitive with CNNs - at least on tasks of intermediate complexity.  In the future, we plan to extend our physics-aware methods to analyse the cross-over points at which various deeper ANNs scale to state-of-the-art tasks given both temporally and physically sequential sub-systems.

\section*{Acknowledgment}
Sandia National Laboratories is a multimission laboratory managed and operated by NTESS, LLC, a wholly owned subsidiary of Honeywell International Inc., for the U.S. Department of Energy’s National Nuclear Security Administration under contract DE-NA0003525. This paper describes objective technical results and analysis. Any subjective views or opinions that might be expressed in the paper do not necessarily represent the views of the U.S. Department of Energy or the United States Government.


\nocite{*} 


\end{document}